# My Brain is Full: When More Memory Helps

**Christopher Lusena** **Tong Li** **Shelia Sittinger** **Chris Wells** **Judy Goldsmith** *

Computer Science Dept.
University of Kentucky

## Abstract

We consider the problem of finding good finite-horizon policies for POMDPs under the expected reward metric. The policies considered are ***free finite-memory policies with limited memory;*** a policy is a mapping from the space of observation-memory pairs to the space of action-memory pairs (the policy updates the memory as it goes), and the number of possible memory states is a parameter of the input to the policy-finding algorithms. The algorithms considered here are preliminary implementations of three search heuristics: local search, simulated annealing, and genetic algorithms. We compare their outcomes to each other and to the optimal policies for each instance. We compare run times of each policy and of a dynamic programming algorithm for POMDPs developed by Hansen that iteratively improves a finite-state controller — the previous state of the art for finite memory policies. The value of the best policy can only improve as the amount of memory increases, up to the amount needed for an optimal finite-memory policy. Our most surprising finding is that more memory helps in another way: given more memory than is needed for an optimal policy, the algorithms are more likely to converge to optimal-valued policies.

## 1 INTRODUCTION

Partially observable Markov decision processes (POMDPs) are mathematical models of controlled systems where the controller is uncertain about both the state of the system and about the effects of actions on the system. While this sounds like a less than optimal state of affairs, it is unfortunately an accurate model of many real systems, from robotic navigation to economics. (For a discussion of applications of POMDPs, see (Cassandra, 1998).)

Slightly more formally, a POMDP is a tuple $M = (\mathcal{S}, \mathcal{A}, \mathcal{O}, t, o, r)$, where $\mathcal{S}$ is a finite set of states, $\mathcal{A}$ is the finite set of actions, $\mathcal{O}$ the finite set of possible observations, $t$ the probabilistic transition function mapping $\mathcal{S} \times \mathcal{A}$ to $\mathcal{S}$, $o$ the observation function mapping $\mathcal{S}$ to $\mathcal{O}$ (deterministically, for this paper), and $r$ the reward function, mapping $\mathcal{S} \times \mathcal{A}$ to the real numbers. Our algorithms require an initial belief state; for the grid worlds, for instance, it is a uniform distribution over all the states.

It is unfortunate that POMDPs model so many real-world situations because it is computationally intractable to find or indeed even to store optimal control policies for general POMDPs. It has been shown that the optimal policies are generally those that take into account the full history of the system; because the number of possible histories increases exponentially with each epoch considered — and is uncountable in the limit — it is infeasible to store policies as tables. Fortunately, some policies can be represented finitely, and sometimes compactly, as functions. However, finding such an optimal policy for a finite horizon is PSPACE-hard (Papadimitriou & Tsitsiklis, 1987); for an infinite horizon, it is uncomputable (Madani, 1998). Furthermore, there can only be a provably good polynomial-time approximation algorithm (a so-called "$\varepsilon$-approximation") if P=PSPACE (Lusena *et al.*, 1998).

Stationary policies, those that take into account only the current observation, are computationally simpler: finding the optimal stationary policy for a POMDP is only NP-hard (Papadimitriou & Tsitsiklis, 1987). However, this is still not considered tractable. Unsurprisingly, there can only be a provably good polynomial-time approximation algorithm if

* {lusena,tongli,smsitt0,chrisw,goldsmit}@cs.uky.edu

NP=P (Lusena *et al.*, 1998). Thus, these policies are unsatisfactory both because of their non-optimality and the difficulty of finding or approximating them.

There is, fortunately, a compromise. In this work, we consider *finite memory* policies. These policies were introduced in Sondik's thesis (Sondik, 1971), but the most general form of finite memory policies has received little attention until recently. (See, for instance, (Hansen, 1998a; Hansen, 1998b; Meuleau *et al.*, 1999).)

Finite memory can be used to record the last $k$ states seen; this restriction, *finite-history policies,* was extensively explored in the '70's and '80's (Lovejoy, 1991). Memory could instead be used to record the time the system has run (for finite horizon problems), yielding *time-dependent policies.* We consider *free finite memory* policies. We allow the policy-finding algorithm to determine the use of the memory without any restriction except a limit on the amount of memory used. This restriction forces the complexity of the problem from PSPACE-hard to NP-hard.

Formally, a free finite memory policy with $\mathcal{M}$ memory states for POMDP $M = (\mathcal{S}, \mathcal{A}, \mathcal{O}, t, o, r)$ is a function $\pi_f : \mathcal{O} \times \mathcal{M} \rightarrow \mathcal{A} \times \mathcal{M}$, mapping each pair ⟨observation, memory state⟩ to a pair ⟨action, memory state⟩, where $\mathcal{M}$ is a finite set of states, i.e., a finite "scratch" memory.

Eric Hansen has done some excellent work on finite-memory policies. Hansen's algorithms (Hansen, 1998a; Hansen, 1998b) represent policies as finite-state controllers: each state of the controller represents an action (not necessarily uniquely), and transitions are made based on the next observation. This is another form of restriction on the use of the finite memory. With finite-state controllers, the transition from one action-state to another depends only on the observation; with free finite memory, the transition may depend on the observation and the memory update.

Hansen's work has a major advantage over the other current work on POMDPs: it is significantly faster. Furthermore, his algorithms, which are probabilistic, usually find good policies. However, we expect that our algorithms will scale significantly better than Hansen's.

The biggest difference between our work and that of others such as Hansen is that we are computing policies *given a predetermined bound on the number of possible memory states.*

Note that a $k$-state free finite memory policy for a POMDP $M = (\mathcal{S}, \mathcal{A}, \mathcal{O}, t, o, r)$ can be considered a stationary policy on a new POMDP, $M'$, where the state space for $M'$ is $\mathcal{S} \times \mathcal{M}$, where $\mathcal{M}$ is the set of $k$ memory states. Thus the complexity results for stationary policies hold for free finite memory policies with limited memory as well.

Our idea of applying search heuristics to the space of free finite memory policies for a POMDP has been applied by Littman (Littman, 1994) and others to the problem of finding good stationary policies. However, in many instances (including those we discuss in Section 6.1) finite-memory policies out-perform the optimal stationary policy significantly. Meuleau et al. have applied search heuristics to finding good finite-state controllers of a fixed size in a learning theoretic context (Meuleau *et al.*, 1999). Their work is fairly similar to ours, but our preliminary results seem to be better, either because we assume knowledge of the model or because we are working with a better update heuristic in our local search algorithms.

A major question that this work begins to explore is, How much finite memory should be used? What we show in Section 6 is that the questions of 'How much memory is optimal for a policy' and 'How much memory is optimal for a search algorithm' may have different answers.

In particular, it seems that giving a search algorithm extra memory to explore greatly increases its probability of finding an optimal policy. Users are thus faced with a common trade-off, since increasing the number of possible memory states greatly increases computation time. **Extra memory states increase the probability of finding a better policy, at a cost of more computation time.**

We should vehemently note that the increased time we discuss is still significantly less than the time needed to compute the optimal history-dependent policy for any but the smallest examples considered, even using the state-of-the-art exact policy finding algorithms such as Incremental Pruning (Zhang & Liu, 1997; Cassandra *et al.*, 1997) or Hansen's Policy Iteration algorithms (Hansen, 1998a; Hansen, 1998b).

The work discussed here is based on three search heuristics: local search (Sec. 2), genetic algorithms (Sec. 3), and simulated annealing (Sec. 4). The resulting policies were compared with the optimal policy for each memory size, as calculated by a branch and bound algorithm. In addition, for each POMDP we calculated a finite-state controller using Hansen's algorithm. We cannot directly compare the finite-state controller to policies found by our algorithms (the controller represents an infinite horizon policy, whereas ours are for finite horizons, and Hansen's algorithms assume a discounted reward whereas ours assume a total reward). We will be considering discounted rewards in the next paper. In the meantime, we show that even running one of our algorithms multiple times and taking the best outcome is often significantly faster than running Hansen's algorithm.

## 2 LOCAL SEARCH

Local search is a technique well-known at least since Biblical times (see Genesis 26:14-33) for finding an optimal value of a function by incremental improvements. A neighborhood structure is defined on the space of solutions, and the algorithm begins with an initial solution. While there is an improved solution in the neighborhood of the current solution, the current solution is updated to some better neighbor. Much of the art of designing good local search algorithms goes into the proper definition of neighborhood and the update heuristic.

### 2.1 RELATED WORK

There are certain parallels between our work and the work done on history-dependent policies, although we find some of the apparent similarities misleading. For instance, both policy and value iteration algorithms are forms of local search: they find local improvements to the policy under consideration, until that is no longer possible. The difference between the computational needs of those algorithms and ours is that our policies tend to be significantly smaller than theirs, so evaluating neighboring policies is faster.

Meuleau, et al., have studied a global branch and bound search algorithm for finding the best deterministic policy graph (similar to our branch and bound algorithm) and a local search algorithm based on gradient descent for finding the best stochastic policy graph. Based on the data in (Meuleau *et al.*, 1999), it appears that this is not as efficient as our local search algorithms, for several reasons. It is set in a learning-theoretic context and it finds finite-state controllers instead of free finite memory policies. We also speculate that they have chosen a less efficient update rule than we have.

### 2.2 OUR ALGORITHM

The update rule used in our algorithm is *randomized first improvement.* For each policy, there is a well-defined ordering on its neighbors. A random initial policy and one of its neighbors are chosen at random. That neighbor is defined to be the first in the list of neighbors (which is wrapped around to include all neighbors). Once that list is defined, the first improvement to the current policy that appears on the list is chosen. This process is repeated until no better neighbor exists, meaning we have found a local maximum.

## 3 GENETIC ALGORITHMS

A genetic algorithm is a randomized search algorithm where there is a "population" of solutions at each time step. The population evolves via mutation, reproduction, and fitness selection. Reproduction involves a combination of the "phenotypes" of two members of the population. This formalism was introduced in (Holland, 1975).

### 3.1 RELATED WORK

Lanzi has used genetic algorithms as a randomizing technique in conjunction with a learning algorithm to learn a particular type of finite-memory policy (Lanzi, 1998a). He considers two extensions to Wilson's XCS system (Wilson, 1995): XCSM (Lanzi, 1998b) and XCSMH (Lanzi, 1998a).

XCS is a Q-learning-like technique which learns a memoryless policy for an environment via separate exploration and exploitation phases. XCSM extends XCS by adding bits of (unconstrained) memory to the policy. Unfortunately, XCSM does not always converge well when there are states with the same observation that require different actions (and thus different memory), since the memory-action pairs are tightly coupled. Therefore, in XCSMH this coupling is loosened in the exploration phase, and additional heuristics are applied.

Thus, XCSMH uses free finite memory; like our algorithms, it uses genetic algorithms to learn a policy. Lanzi's work, however, uses an additional constraint and additional heuristics because he is solving a learning-theoretic problem rather than a control-theoretic one. Therefore it is difficult to compare the outcomes of these heuristics.

Lin, et al., use genetic algorithms to construct sets of vectors that approximate the value function for the $k$-step finite horizon discounted history-dependent policy (Lin *et al.*, 1998). However, they are solving a very different problem from ours.

### 3.2 OUR ALGORITHM

In our model, a policy corresponds to a phenotype with one "chromosome." A decision for a given observation-memory pair is a single "gene," and a particular decision (action-memory pair) is an "allele" for that gene.

For a particular POMDP, a population size must first be determined. A small population will prevent diversity and produce poor policies, while a large population will cause the algorithm to take too long. The population size should therefore be at least partially dependent on the size of the policy space. We arbitrar-ily set 30 as the smallest population size, and make all populations grow with the log of the size of the policy space, since it is this space in which we are searching. Because we expect more memory to deliver better policies, however, the size of the policy space is determined under the assumption that no memory will be used. So our population grows linearly with the number of observations in the POMDP, and logarithmically with the actions and with the number of memory states.

The population is filled initially at random. During a generation, the fitness of each phenotype (policy) is determined by evaluating the policy over a finite horizon. The values of each policy are used to determine the policies' fitnesses. Any policy below two standard deviations of the population's mean of fitness is discarded. Any policy above two standard deviations of the mean has its fitness set to two standard deviations above the mean, in order to prevent a very good policy from destroying the diversity of the matring pool. We thus eliminate all outliers. Once the fitness of each policy has been determined, policies are selected based on their fitness to fill the new mating pool.

The mating pool is altered according to the cross-over rate and the mutation rate. We have not yet studied the effects of different rates on generated policies, so we use a preliminary cross-over rate of .5, and a preliminary mutation rate of .005. The number of pairs that are crossed over is equal to the population size times the cross-over rate. A random observation and a random memory state are selected. All decisions above the selected observation and memory state, and all decisions below the selected observation and memory state, are swapped between the two policies. No two policies are selected more than once for cross-over per generation. After cross-over, $p \cdot m$ policies are selected for mutation, where $p$ is the population size and $m$ is the mutation rate. In a mutation, a random decision in a policy is selected, and has its action and memory state randomly changed to a new action/memory-state pair.

Termination is determined during selection. If the same best policy occurs more than a set number of times (we use a value of 10), or the best policy occurs half this number of times and the standard deviation of the population drops below a given threshold (we use a threshold of .0001), improvement is not taking place. The genetic algorithm halts under either of these conditions.

# 4 SIMULATED ANNEALING

Simulated annealing is another heuristic for improving local search methods. It was introduced by Kirkpatrick, et al. (Kirkpatrick *et al.*, 1983). The method is applied by selecting a starting "temperature" - the higher the temperature, the more frequently a spontaneous change of state occurs in the otherwise local search. As the changes become smaller, the temperature is reduced, and the search becomes a purely local search. The expectation is that the initial high rate of change will help prevent convergence to a suboptimal local optimum.

## 4.1 RELATED WORK

There are many techniques for approximating the value of a state under a given history-dependent policy. Some of those use randomized techniques either implicitly, in learning algorithms for example, or explicitly. (See (Lin *et al.*, 1998) for an example of simulated annealing applied to this problem.)

## 4.2 OUR ALGORITHM

In this experiment, the initial temperature was chosen as 95, giving a 95% probability of change at each initial local search test. The temperature was decreased by one degree at each iteration. This process was continued until a local maximum was achieved, and the temperature dropped to 0.

# 5 FINITE-STATE CONTROLLERS

A finite-state controller is a finite state automaton representing a policy. The states of the automaton correspond to actions and the transitions of the automaton correspond to observations. There is a well-understood correspondence between vectors of the value function as constructed in value iteration on history-dependent policies for POMDPs and states of a finite controller (Hansen, 1998a).

The difference between a finite-state controller and a free finite memory policy is that for a finite-state controller, there is a tight link between the states of the controller (the policy's memory) and action, whereas free finite memory policies can decouple this link.

## 5.1 RELATED WORK

Hansen has several algorithms for finding finite-state controllers for POMDPs. Although there are some significant differences between his work and ours (infinite-horizon vs. finite-horizon, controllers vs. free finite memory), we have implemented his policy iteration algorithm for comparison purposes.

In Hansen's policy iteration algorithm (Hansen, 1998a), a policy is represented by a finite-state controller. Finding an $\varepsilon$-optimal policy for infinite hori-zon POMDPs is done by iteratively improving a finite-state controller, where each machine state corresponds to a vector of the value function and is associated with an action choice. Machine state transitions are labeled by observations. Explicitly representing a policy as a finite-state controller makes the policy evaluation very easy. Policy improvement involves adding, changing, and pruning machine states. Because it improves the value function both by the dynamic programming update and by policy evaluation, policy iteration outperforms value iteration greatly.

Meuleau et al. present another approach for solving POMDPs by searching in the policy space (Meuleau *et al.*, 1999). In their work, policies are also represented by finite-state controllers (policy graphs is the term they use). Like us, they consider *a priori* constraints on the amount of memory used; unlike us, their memory is restricted to the form of a finite controller. Their evaluation algorithm is based on an MDP defined by the cross-product of the POMDP and a policy graph; this is much faster than the evaluation algorithms used in our work. Since they impose extra constraints on the policy graph, the policy search space is reduced, so large POMDPs can be solved significantly more quickly than with classical approaches.

There are other restrictions on finite memory. For instance, in (Wiering & Schmidhuber, 1997), Weiring and Schmidhuber introduce a notion of HQ-learning that breaks down goal POMDPs into separate subproblems, each of which is solved by a separate agent. These agents can be thought of as corresponding to different memory states. However, this decomposition does not directly correspond to our finite memory policies, since their overall policy is limited to using each agent once, in sequence.

### 5.2 COMPARISONS WITH OUR WORK

We have implemented Hansen's policy iteration algorithm, and compared our search techniques with it. In those cases where his algorithm halted in less than two hours, the time was comparable, or in some cases better, than most of our algorithms. However, for all the larger instances of POMDPs (for instance, mccallum, sutton, aloha10 and aloha30) the process either continued past our two-hour cut-off (in the case of mccallum, we stopped it at approximately 2,407 minutes) or died due to numerical instabilities [1].

[1]Hansen ran 4x3.95 in his dissertation. It stopped after 10,681 seconds. It is not clear that our implementation would do so, since he used a superior LP programming package, CPLEX, and some instance-tuning of parameters.

Figure 1: Average Run Times on aloha.10

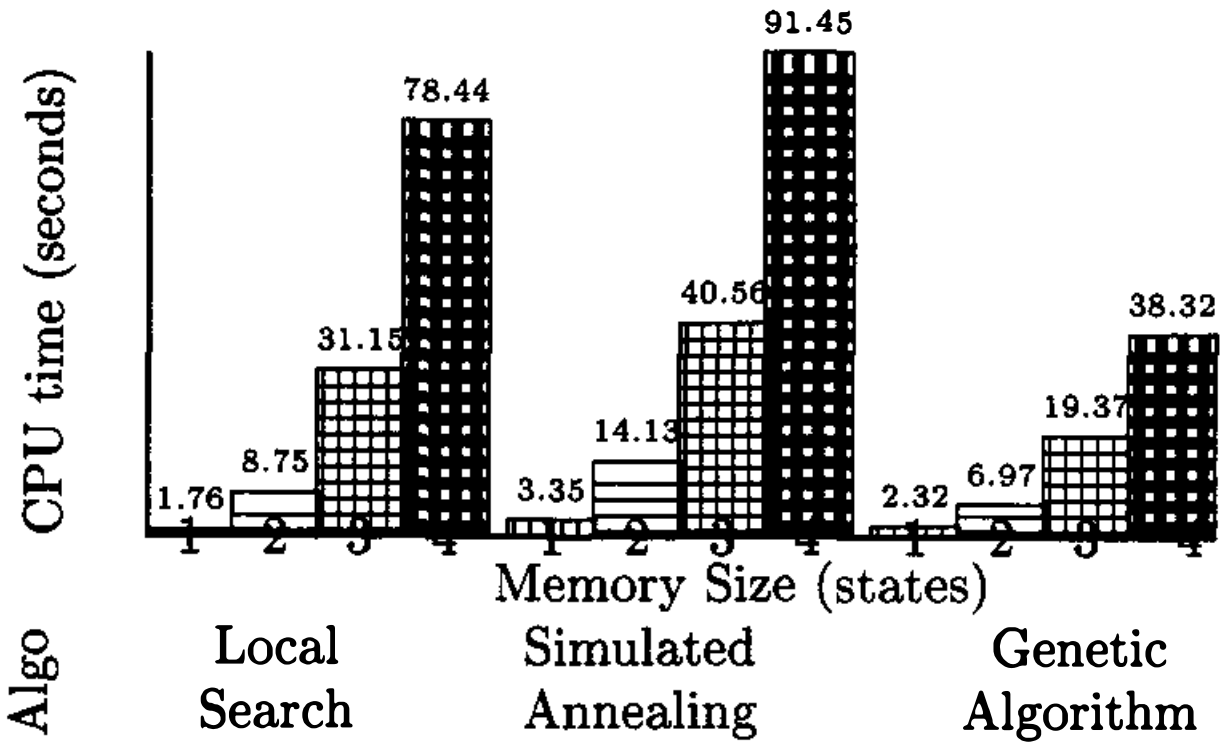

## 6 RESULTS

### 6.1 EXAMPLE POMDPS

We have run our algorithms on ten POMDPs (five from Cassandra's database (Cassandra, 1997-9), four others from the literature, and an additional small test case: 1D, 4x3.95, 4x3.CO.95, aloha.10, aloha.30; Sutton's gridworld and McCallum's maze (Littman, 1994), Maze7 and Maze10 (Lanzi, 1998a), and test1), for 1, 2, 3, and 4 memory states. In addition, we considered 1-10 memory states for 1D and 1-7 memory states for Maze7 and Maze10. We ran each algorithm 100 times on each input and averaged the results where applicable.

### 6.2 TIME

When we consider the run time graphs for all our examples, we discover that the curves from the different POMDPs do not match. However, for each instance we got results similar to those shown in Figure 1 for `aloha.10`. However, usually the genetic algorithm ran more slowly, on average, than local search or simulated annealing.

Of the 10 POMDPs considered here, 7 can be characterized as *goal oriented*. One measure of computational effort is to count the number of policy evaluations for each instance of local search. In Figure 2, we plot the average number of evaluations of policies versus the log of the size of the policy space. The plotted line is a quartic polynomial fitted to the data using GNUPLOT, in order to provide a sense of the relationship between the size of the POMDP and the time taken to solve it. (The error bars indicate one standard deviation from the mean in all the time graphs.) Notice that, because of the restrictions placed on the amount of memory used, the size of the policy space is bounded by $2^{O(n \log n)}$, where $n$ is the size of the input, so the log of the size of the policy space is polynomial

Figure 2: Local Search on Goal Oriented POMDPs

Number of Evaluation vs. Policy Space

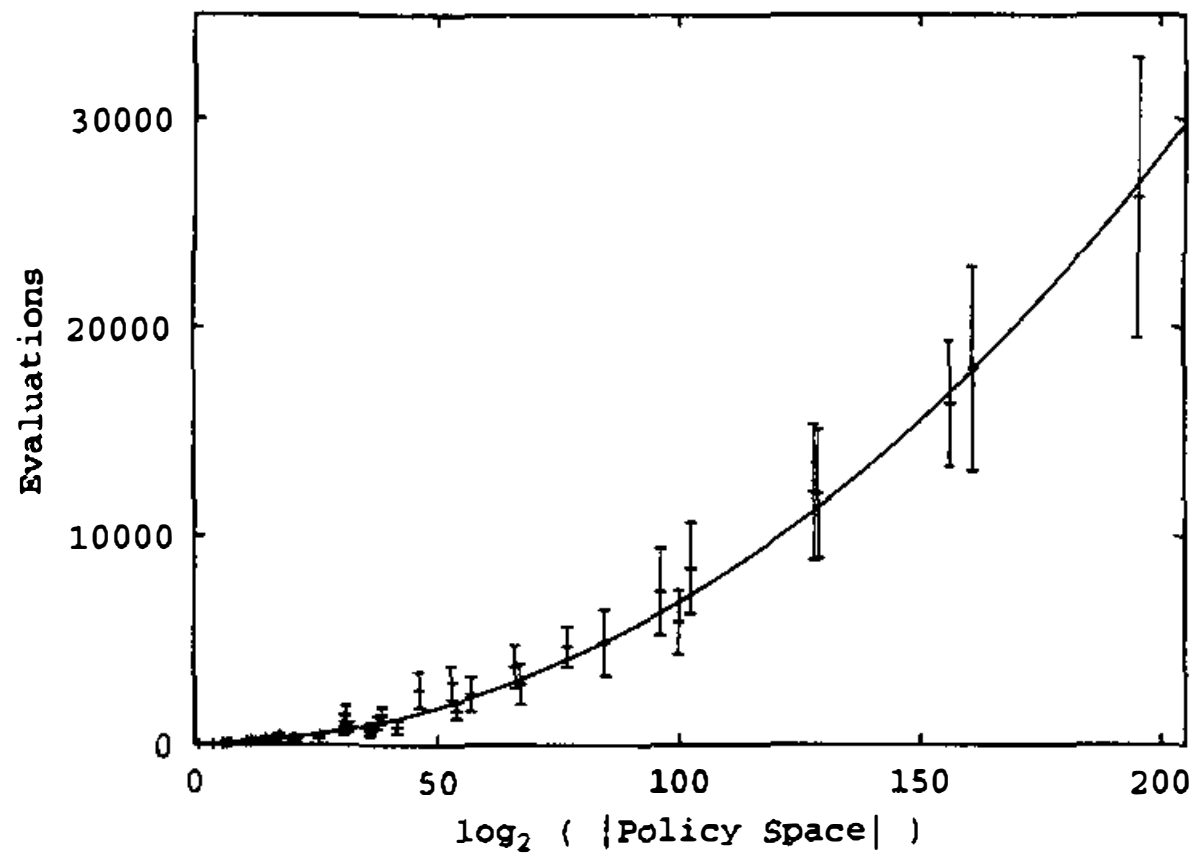

Figure 3: Simulated Annealing on Goal Oriented POMDPs

Number of Evaluation vs. Policy Space

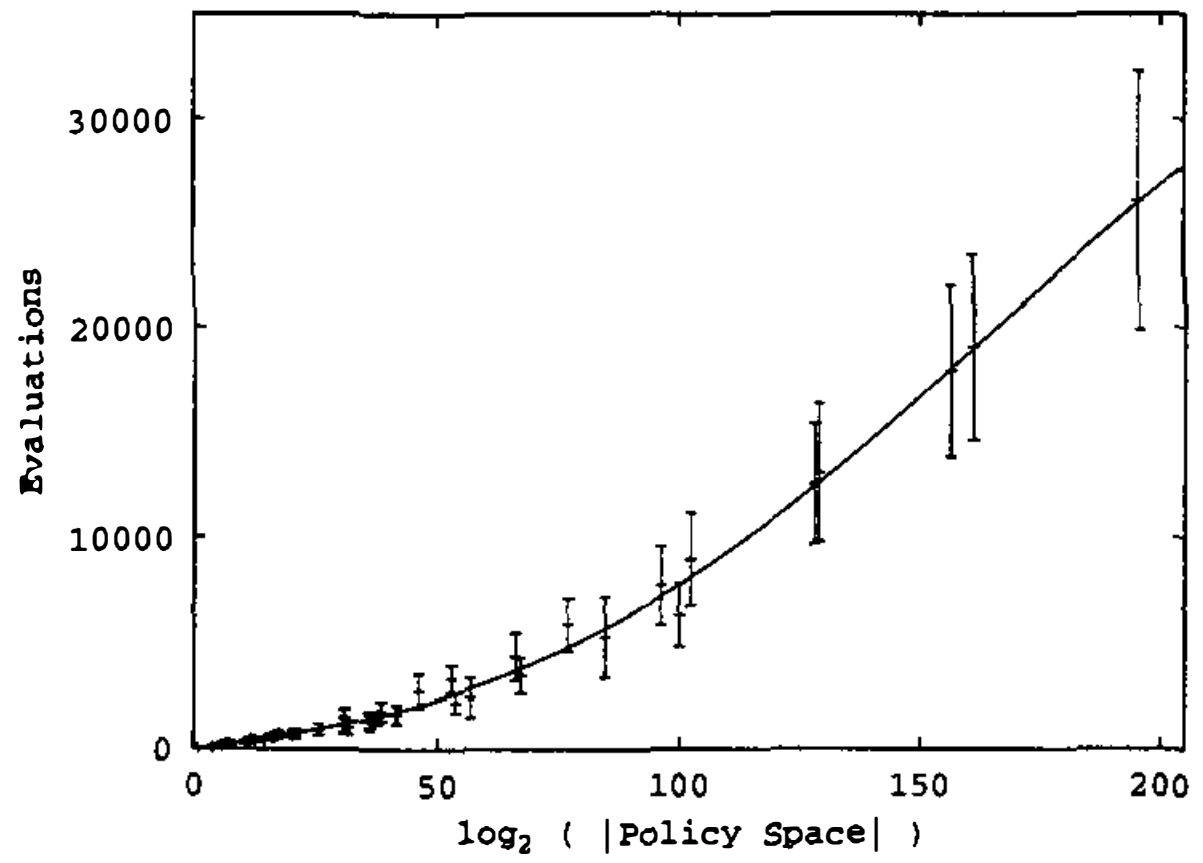

in $n$.

Based on all of our runs, local search appears to run in time polynomial in the size of the input, on average over the examples we have tested.

As we see in Figure 3, simulated annealing requires roughly the same number of evaluations as local search. Since simulated annealing performs slightly better than local search in most instances, the slight increase in time seems worth it.

On the other hand, with our current genetic algorithm, there is no tightly fitted curve for our data with more than one POMDP.

Unsurprisingly, local search is faster than either genetic algorithms or simulated annealing.

### 6.3 EXPECTATION OF FINDING THE OPTIMAL POLICY

There is no consistent predictor of which search algorithm is most likely to find the optimal policy. In our current implementations, the genetic algorithm is the least likely for most POMDPs; we expect to improve that implementation substantially in upcoming experiments. Figure 4 shows our results.

### 6.4 MORE MEMORY IS BETTER!

The most striking result of our research is that searches run with extra memory, i.e., more memory than is needed to get an optimal policy, are still more likely to actually *find* an optimal policy than searches that are constrained to policies with no "extra" memory. For instance, McCallum's maze has an optimal policy with only two memory states, yet Figure 5 shows that the optimal value is found more often when the algorithms are run with more memory.

One explanation of this phenomenon is geometric: if there is a local optimum between the starting policy and the global optimum and only one path from the start to the global optimum, a local search will get stuck at the local optimum. However, the extra memory functions as an additional dimension, allowing multiple paths from the initial policy to the global optimum. (Imagine trying to do hill climbing on the curve $x^3 - x$ from the point $(-1, 0)$: in two dimensions, one gets stuck on the hump at $x = -1/3$, but in three dimensions one might get a surface which does not peak at that $x$ value for all $z$ values, so one might be able to work around that local hump.)

The flip side of this discovery is that adding memory slows down computation, since it increases the size of the policy space. Therefore, one is faced with the time-quality trade-off: allowing the search algorithm to search a larger space of policies dramatically increases the probability of finding an optimal policy, but will take longer to converge.

This effect was consistent for all the POMDPs considered and for all three search algorithms. We demonstrate it here in Figure 5.

## 7 ONGOING WORK

Several extensions and improvements can and should be made to the preliminary implementations used in this work. In (Wells *et al.*, 1999) the positions of the cells in the policies are made maleable, enabling the use of the inversion genetic operator. Additionally, the values of the genetic parameters (such as the crossover and mutation rates) are studied in relation to the quality of generated policies and the speed with which they are produced. Similarly, a better cooling scheme in simulated annealing could lead to converging more quickly to better policies. We are also experimenting with other search space explorations in local search. In a submitted paper, (Lusena *et al.*, 1999), we describe an extension to this work in which we evaluate policies with infinite horizons and discounted rewards.

Figure 4: Likelihood of Finding the Optimal Policy

| POMDP File | Memory Size | Local Search | Simu. Ann. | Genet. Alg. |
|---|---|---|---|---|
| 1d | 1 | 100% | 100% | 100% |
| 1d | 2 | 31% | 28% | 78% |
| 1d | 3 | 25% | 14% | 24% |
| 1d | 4 | 35% | 40% | 33% |
| 1d | 5 | 43% | 57% | 38% |
| 1d | 6 | 63% | 64% | 35% |
| 1d | 7 | 58% | 71% | 32% |
| 1d | 8 | 63% | 70% | 42% |
| 1d | 9 | 66% | 68% | 33% |
| 1d | 10 | 71% | 66% | 24% |
| test1 | 1 | 100% | 100% | 100% |
| test1 | 2 | 18% | 33% | 74% |
| test1 | 3 | 40% | 53% | 96% |
| test1 | 4 | 57% | 61% | 99% |
| mccallum | 1 | 100% | 100% | 40% |
| mccallum | 2 | 34% | 72% | 36% |
| mccallum | 3 | 57% | 87% | 30% |
| mccallum | 4 | 63% | 90% | 32% |
| 4x3.95 | 1 | 100% | 100% | 64% |
| 4x3.95 | 2 | 12% | 24% | 0% |
| 4x3.95 | 3 | 17% | 13% | 0% |
| 4x3.95 | 4 | 3% | 4% | 0% |
| 4x3.CO.95 | 1 | 100% | 100% | 73% |
| maze7 | 1 | 100% | 100% | 24% |
| maze7 | 2 | 42% | 39% | 10% |
| maze7 | 3 | 68% | 47% | 10% |
| maze7 | 4 | 71% | 46% | 4% |
| maze7 | 5 | 83% | 68% | 4% |
| maze7 | 6 | 80% | 73% | 3% |
| maze7 | 7 | 84% | 78% | 1% |
| maze10 | 1 | 66% | 85% | 17% |
| maze10 | 2 | 6% | 4% | 0% |
| maze10 | 3 | 1% | 6% | 0% |
| maze10 | 4 | 3% | 1% | 0% |
| maze10 | 5 | 5% | 4% | 0% |
| maze10 | 6 | 7% | 1% | 0% |
| maze10 | 7 | 5% | 1% | 0% |
| sutton | 1 | 44% | 59% | 24% |
| sutton | 2 | 8% | 4% | 1% |
| sutton | 3 | 16% | 9% | 0% |
| sutton | 4 | 10% | 8% | 0% |
| aloha.10 | 1 | 100% | 100% | 50% |
| aloha.10 | 2 | 61% | 62% | 0% |

Figure 5: Policies Found (McCallum's Maze)

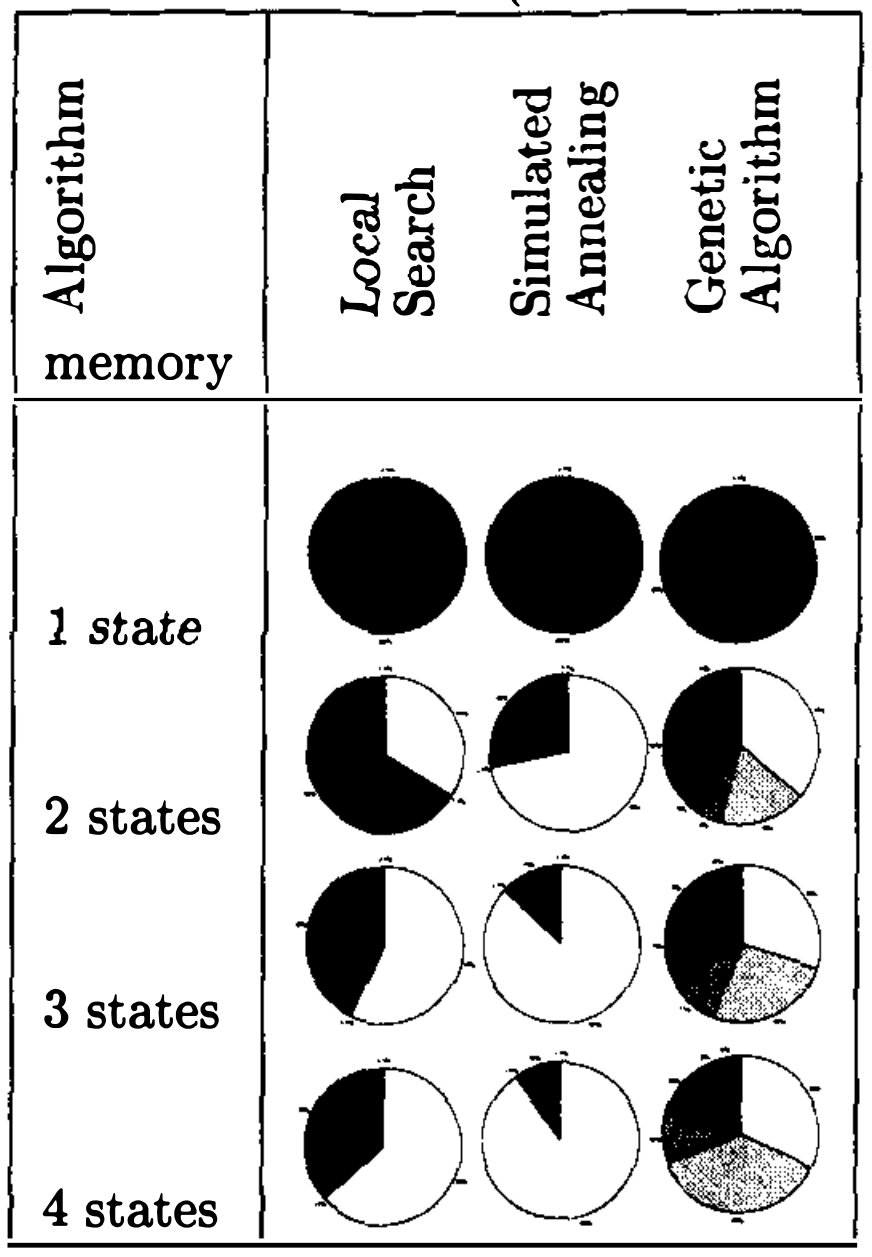

| Colour | Policy Value |
|---|---|
| (black) | < 7.63 |
| (black) | 7.63 (Optimal for 1 Memory State) |
| (black) | > 7.63 ≤ 9 |
| (dark grey) | > 9 ≤ 10 |
| (light grey) | > 10 < 11.27 |
| (white) | 11.27 (Optimal for 2+ Memory States) |

## Acknowledgements

This research supported in part by NSF grant CCR-9610348. The authors would like to thank Eric Hansen for pointers to relevant literature, Rafi Finkel and Beth Goldstein for their guide through biblical searches, and Emily Hendren for her help in editing.